\documentclass[final]{cvpr}

\usepackage{times}
\usepackage{epsfig}
\usepackage{graphicx}
\usepackage{amsmath}
\usepackage{amssymb}
\usepackage{multirow}
\usepackage{pifont}
\newcommand{\cmark}{\ding{51}}%
\newcommand{\xmark}{\ding{55}}%

\newcommand\blfootnote[1]{%
  \begingroup
  \renewcommand\thefootnote{}\footnote{#1}%
  \addtocounter{footnote}{-1}%
  \endgroup
}

\usepackage[pagebackref=true,breaklinks=true,colorlinks,bookmarks=false]{hyperref}

\def\cvprPaperID{24} 
\def\confYear{CVPR 2021}

\begin{document}

\title{Cross-modal Speaker Verification and Recognition: A Multilingual Perspective}

\author{Shah Nawaz$^{1*}$, Muhammad Saad Saeed$^{4*}$, Pietro Morerio$^{1}$, Arif Mahmood $^{5}$, Ignazio Gallo$^{3}$, \\ Muhammad Haroon Yousaf$^{4}$, Alessio {Del Bue}$^{1,2}$  \\
$^{1}$Pattern Analysis \& Computer Vision (PAVIS) - Istituto Italiano di Tecnologia (IIT),\\ 
$^{2}$Visual Geometry \& Modelling (VGM) - Istituto Italiano di Tecnologia (IIT), \\ $^{3}$University of Insubria,\\ 
$^{4}$University of Engineering and Technology Taxila,
$^{5}$Information Technology University 

\\
{\tt\small \{shah.nawaz,pietro.morerio,alessio.delbue\}@iit.it,}
{\tt\small  arif.mahmood@itu.edu.pk},\\
{\tt\small \{haroon.yousaf,saad.saeed\}@uettaxila.edu.pk,}{\tt\small  ignazio.gallo@uninsubria.it}\\

}

\maketitle

\begin{abstract}
Recent years have seen a surge in finding association between faces and voices within a cross-modal biometric application along with speaker recognition. Inspired from this, we introduce a challenging task in establishing association between faces and voices across multiple languages spoken by the same set of persons. The aim of this paper is to answer two closely related questions: \textit{``Is face-voice association language independent?''} and \textit{``Can a speaker be recognized irrespective of the spoken language?''}. These two questions are important to understand effectiveness and to boost development of multilingual biometric systems. To answer these, we collected a Multilingual Audio-Visual dataset, containing human speech clips of $154$ identities with $3$ language annotations extracted from various videos uploaded online. Extensive experiments on the two splits of the proposed dataset have been performed to investigate and answer these novel research questions that clearly point out the relevance of the multilingual problem.
\end{abstract}

\section{Introduction}

\begin{figure}[!t]
    \centering
    \includegraphics[width=0.95\linewidth]{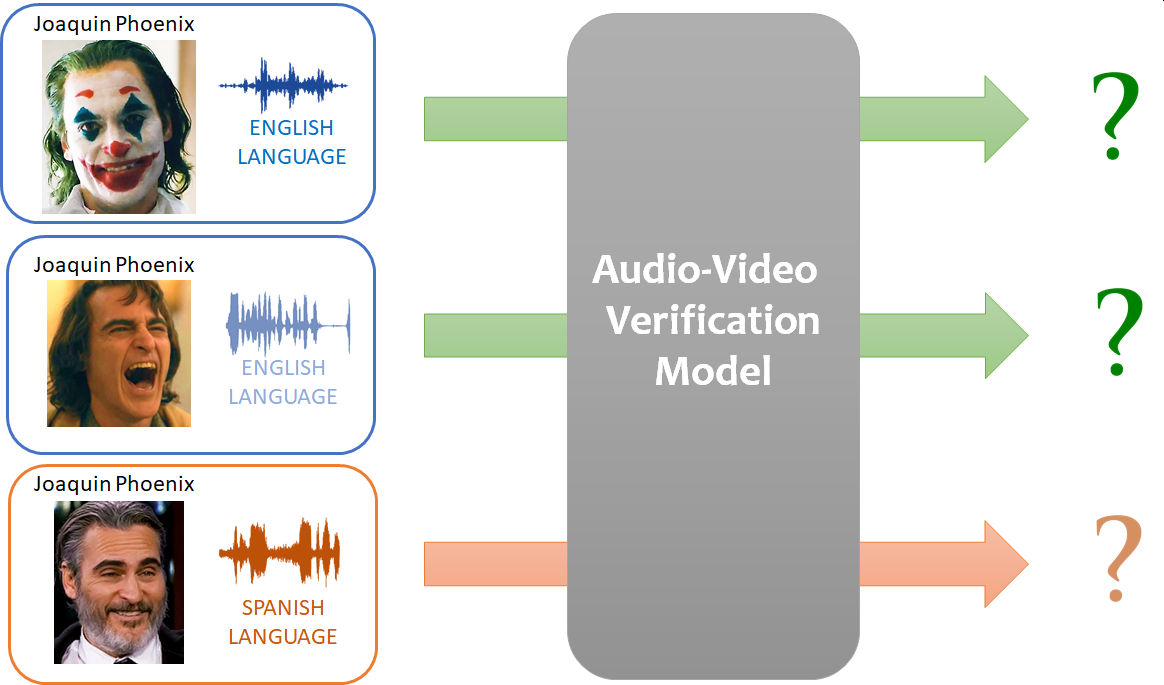}
    \caption{Multimodal data may provide enriched understanding to improve verification performance. Joaquin can wear make-up that makes visual identification challenging but voice can still bring enough cues to verify  identity.  In this work, we are interested to understand the effect of multilingual input when processed by audio-visual verification model (Q1) or just using the audio input (Q2). Joaquin is a perfect English-Spanish bilingual, would the system still be able to verify Joaquin when speaking Spanish even if the system was trained with English audio only?}
    \label{fig:intro}
 \end{figure}

Half of the world population is bilingual with people often  switching between their first and second language while communicating~\cite{bworld}. 
Therefore it is essential to investigate the effect of multiple languages on computer vision and machine learning tasks.  
\blfootnote{*Equal contribution}
As introduced in Fig.~\ref{fig:intro}, this paper probes two closely related questions, which deal with the recent introduction of cross-modal biometric matching tasks in the wild: \\ \newline
Q1. Is face-voice association language independent? \\
Q2. Can a speaker be recognised irrespective of the spoken language? \\ \newline

Regarding the first question, a strong correlation has been recently found between face and voice of a person which has attracted significant research interest~\cite{horiguchi2018face,kim2018learning,nagrani2018learnable,nagrani2018seeing,nawaz2019deep,wen2021seeking,wen2018disjoint}. Though previous works have established an association between faces and voices, however none of these approaches investigate the effect of multiple languages on this task. 
%
In addition, existing datasets containing audio-visual information, \textit{VoxCeleb}~\cite{Nagrani19,Chung18,Nagrani17}, \textit{FVCeleb}~\cite{horiguchi2018face}, \textit{FVMatching}~\cite{kim2018learning} do not provide language level annotation.  Therefore, we cannot deploy these datasets to analyse the effect of multiple languages on association between faces and voices. 

Thus, in order to answer both questions, we create a new Multilingual Audio-Visual \textit{MAV-Celeb} dataset comprising of video and audio recordings with a large number of celebrities speaking more than one language in the wild. The proposed dataset paves the way to analyze the impact of multiple languages on association between faces and voices. Then, we propose a cross-modal verification approach to answer Q1 by analyzing the effect of multiple languages on face-voice association.
In addition, the audio part of the dataset supplies samples of $3$ languages with annotations which serves as a foundation to answer Q2 with speaker recognition baselines. \\
To summarise the paper main contributions are listed as follows:
\begin{itemize}
\itemsep0em 
\item [--] We first propose a cross-modal verification approach to analyze the effect of multiple languages on face-voice association; 
\item [--] Likewise, we perform an analysis that highlights the very same problem  of multilingualism for speaker recognition;
\item [--] We propose the \textit{MAV-Celeb} dataset, containing $2,182$ language-annotated human speech clips with $41,674$ utterances of $154$ celebrities, extracted from videos uploaded online.
\end{itemize}

The rest of the paper is structured as follows: Section~\ref{sec:related-work} explores the related literature on the two introduced questions along with existing datasets. While Section~\ref{sec:data-des} introduces the nature of proposed dataset followed by experimental evidence to answer both   questions in Section~\ref{sec:first} and~\ref{sec:second}. Finally, conclusion is presented in Section~\ref{sec:con}.


\setlength{\tabcolsep}{4pt}
\begin{table*}[t]
\begin{center}
\begin{tabular}{|lccc|}
\hline
Dataset & Condition &  Free & Language annotations  \\
\hline\hline
The Mixer Corpus~\cite{cieri2004mixer}       & Telephone, Microphone   & \xmark   & \cmark \\
Vermobil~\cite{burger2000verbmobil}          & Telephone, Microphone   & \xmark   & \cmark \\
Call My Net Corpus~\cite{jones2017call}      & Telephone               & \cmark   & \cmark \\
Common Voice~\cite{ardila2019common}         & Microphone              & \cmark   & \cmark \\
SITW~\cite{mclaren2016speakers}              & Multimedia              & \cmark   & \xmark \\
VoxCeleb~\cite{Nagrani19,Chung18,Nagrani17}  & Multimedia              & \cmark   & \xmark  \\
MAV-Celeb (Proposed)                         & Multimedia              & \cmark   & \cmark  \\
\hline
\end{tabular}
\end{center}
\caption{Comparison of the proposed dataset with existing datasets.}
\label{tab:existing-datasets}
\end{table*}
\setlength{\tabcolsep}{1.4pt}


\setlength{\tabcolsep}{4pt}
\begin{table*}[t]
\begin{center}
\begin{tabular}{|lcc|}
\hline
Dataset & EU & EH  \\
\hline\hline
Languages                           & U/E/EU           & H/E/EH           \\
\# of Celebrities                   & 70               & 84               \\
\# of male celebrities              & 43               & 56              \\
\# of female celebrities            & 27               & 28              \\
\# of videos                        & 560/406/966      & 546/668/1214    \\
\# of hours                         & 59/32/91         & 48/60/109       \\
\# of utterances                    & 11835/6550/18385 & 9974/13313/23287 \\
Avg \# of videos per celebrity      & 8/6/14           & 6/8/14           \\
Avg \# of utterances per celebrity  & 169/94/263       & 119/158/277      \\
Avg length of utterances(s)         & 17.9/17.8/17.8   & 17.4/16.5/16.9   \\
\hline
\end{tabular}
\end{center}
\caption{Dataset statistics. The dataset is divided into $2$ splits (EU, EH) containing audio samples from $3$ languages, English(E), Hindi(H) and Urdu (U).}
\label{tab:datasaet}
\end{table*}
\setlength{\tabcolsep}{1.4pt}

\section{Related Work}
\label{sec:related-work}
We summarize previous work relevant to the two questions raised in the introduction. Q1 falls under cross-modal verification topic  while Q2 deals  with speaker recognition task.

\subsection{Cross-modal Verification Between Faces and Voices}
Last decade has witnessed an increasing use of multimodal data in challenging Computer Vision tasks including visual question and answering~\cite{anderson2018bottom,antol2015vqa}, image captioning~\cite{karpathy2015deep,vinyals2015show}, classification~\cite{gallo2017multimodal,kiela2018efficient}, cross-modal retrieval~\cite{nawaz2019cross,wang2016learning} and multimodal named entity recognition~\cite{arshad2019aiding,zhang2018adaptive}.
Typically, multimodal applications are built on image and text information, however recent years have seen an increased interest to leverage audio-visual information~\cite{huang2013audio,ngiam2011multimodal,srivastava2012multimodal,wu2019dual}. Previous works~\cite{albanie2018emotion,aytar2016soundnet} capitalize on natural synchronization between audio and visual information to learn rich audio representation via cross-modal distillation.
More recently, Nagrani et al.~\cite{nagrani2018seeing} leveraged audio and visual information to establish an association between faces and voices in a cross-modal biometric matching. 
Furthermore, recent works~\cite{kim2018learning,nagrani2018learnable} introduced joint embedding to establish correspondences between faces and voices. These methods extract audio and face embedding to minimize the distance between embeddings of similar speakers while maximizing the distance among embeddings from different speakers.
The framework used speaker identity information to eliminate the need of pairwise or triplet supervision~\cite{nagrani2018learnable,nagrani2018seeing}. 
Wen et al.~\cite{wen2018disjoint} presents a disjoint mapping network to learn a shared representation for audio and visual information by mapping them individually to common covariates (gender, nationality, identity). Similarly, Nawaz et al.~\cite{nawaz2019deep} extracted audio and visual information with a single stream network to learn a shared deep latent space representation.

Our goal is similar to previous works~\cite{kim2018learning,nagrani2018learnable,nagrani2018seeing,wen2018disjoint,nawaz2019cross}, however, we investigate a novel problem: To understand if the association between faces and voices is language independent. 

\subsection{Speaker Recognition}
Speaker recognition dates back to 1960s when Sandra et al.~\cite{pruzansky1963pattern} laid the groundwork for speaker recognition systems attempting to find a similarity measure between two speech signals by using filter banks and digital spectrograms. In the following we provide a brief overview of speaker recognition methods as clustered in two main classes: Traditional and deep learning methods.\\
\textbf{Traditional Methods~--~} For a long time, low dimensional short-term representation of audio input has been basis for speaker recognition tasks e.g. Mel Frequency Cepstrum Coefficients (MFCC) and Linear Predictive Coding (LPC) based features. These features are extracted using short overlapping segments of audio samples. 
Reynolds et al.~\cite{reynolds2000speaker} introduced speaker verification method based on Gaussian Mixture Models using MFCCs. 
Differently, Joint Factor Analysis (JFA) models speaker and channel subspace separately~\cite{kenny2005joint}. Najim et al.~\cite{dehak2009support} introduced i-vectors which combines both JFA and Support Vector Machines (SVM). Other works employed JFA as a feature extractor in order to train a SVM classifier. Furthermore, traditional methods have also been applied to analyze the effect of multiple languages on speaker recognition tasks~\cite{auckenthaler2001language,lu2009effect,misra2014spoken}. Though, traditional methods showed reasonable performance on speaker recognition task, however these methods suffer performance degradation in real-world scenarios. \\
\textbf{Deep Learning Methods~--~}Neural Networks have provided  more efficient methods of speaker recognition. Therefore, the community has experienced a shift from hand-crafted features to deep neural networks.  Ellis et al.~\cite{ellis2001tandem} introduced a system in which a Gaussian Mixture Model is trained from  embedding of hidden layers of a neural network.
Salman et al.~\cite{salman2011exploring} proposed a deep neural network which learn from speaker-specific characteristics from MFCC features for segmentation and clustering of speakers. 
Chen et.al.~\cite{chen2011residual} used a Siamese feed forward neural network which can discriminatively compare two voices based on MFCC features. Lei et al.~\cite{lei2014novel} introduced a deep neural model with i-vectors as input features for the task of automatic speaker recognition. Nagrani et al.~\cite{Nagrani17} proposed adapted convolutional neural network (VGG-Vox) which aggregate frame-level feature vectors to obtain a fixed length utterance-level embedding. More recently, Xie et al.~\cite{xie2019utterance} improved this frame-level aggregation with NetVLAD or GhostVLAD layer.  
This paper has similarities with the previous work i.e. speaker identification and verification, however the objective is different: We evaluate and provide an answer about the effect of multiple languages on speaker identification and verification strategies in the wild. To this end we propose a dataset instrumental for answering such questions.

\subsection{Related Datasets}
There are various existing datasets for multilingual speaker recognition task but they are not instrumental to answer Q1/Q2 due to at least one of the following reasons: i) they are obtained in constrained environment~\cite{cieri2004mixer}; ii) they are manually annotated so limited in size; iii) not freely available~\cite{burger2000verbmobil};  iv) not audio-visual~\cite{cieri2004mixer,jones2017call} v) missing language annotations \cite{Nagrani19,Chung18,Nagrani17}. A comparison of these dataset with our proposed MAV-Celeb dataset is provided in Table~\ref{tab:existing-datasets}.

\section{Dataset Description}
\label{sec:data-des}
Multilingual Audio-Visual~\textit{MAV-Celeb} dataset provide data of $154$ celebrities in $3$ languages (English, Hindi, Urdu). 
These three languages have been selected because of several factors: i) They represent approximately 1.4 Billion bilingual/trilingual people; ii) The population is highly proficient in two or more languages; iii) There is a relevant corpus of different media that can be extracted from available online repositories (e.g. YouTube).
The collected videos cover a wide range of unconstrained, challenging multi-speaker environment including political debates, press conferences, outdoor interviews, quiet studio interviews, drama and movie clips.

 \begin{figure*}[t]
     \centering
     \includegraphics[scale=0.20]{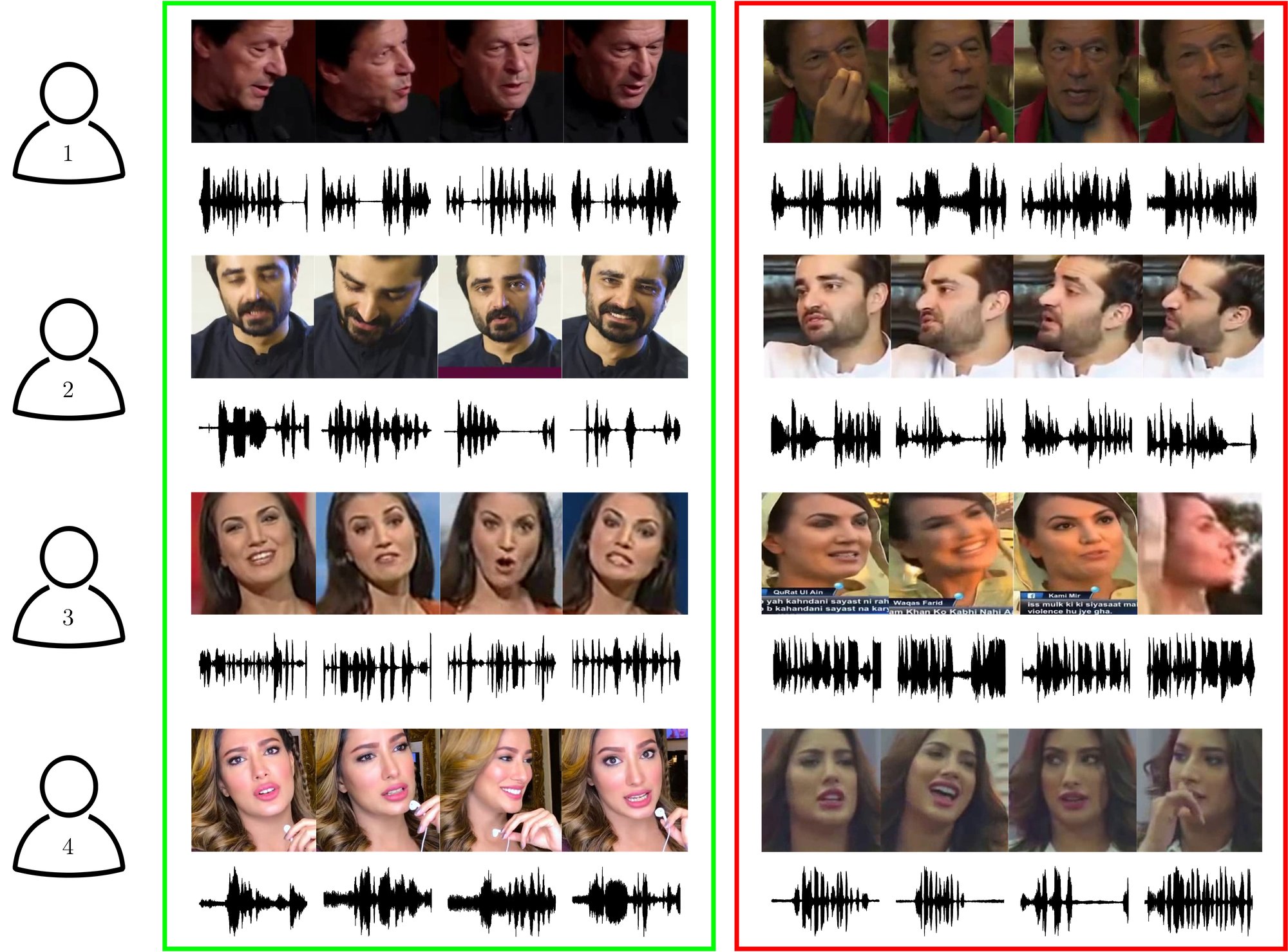}
     \caption{Audio-visual samples selected from proposed dataset. The visual data contains various variations such as pose, lighting condition and motion. The~\textcolor{green}{\textbf{green}} block contains information of celebrities speaking English and the~\textcolor{red}{\textbf{red}} block presents data of the same celebrity in Urdu.} 
     \label{fig:cropped_faces}
 \end{figure*}

It is also interesting to note that the visual data spans over a vast range of variations including poses, motion blur, background clutter, video quality, occlusions and lighting conditions. In addition, videos are degraded with real-world noise like background chatter, music, overlapping speech and compression artifacts. 
Fig.~\ref{fig:cropped_faces} shows some audio-visual samples while Table~\ref{tab:datasaet} shows statistics of the dataset. The dataset contains $2$ splits English--Urdu (EU) and English--Hindi (EH) to analyze  performance measure across multiple languages. 
The pipeline followed in creating the dataset is discussed in Appendix~\ref{append:data-pipeline}. 

 \begin{figure*}[t]
 \centering
 \includegraphics[scale=0.055]{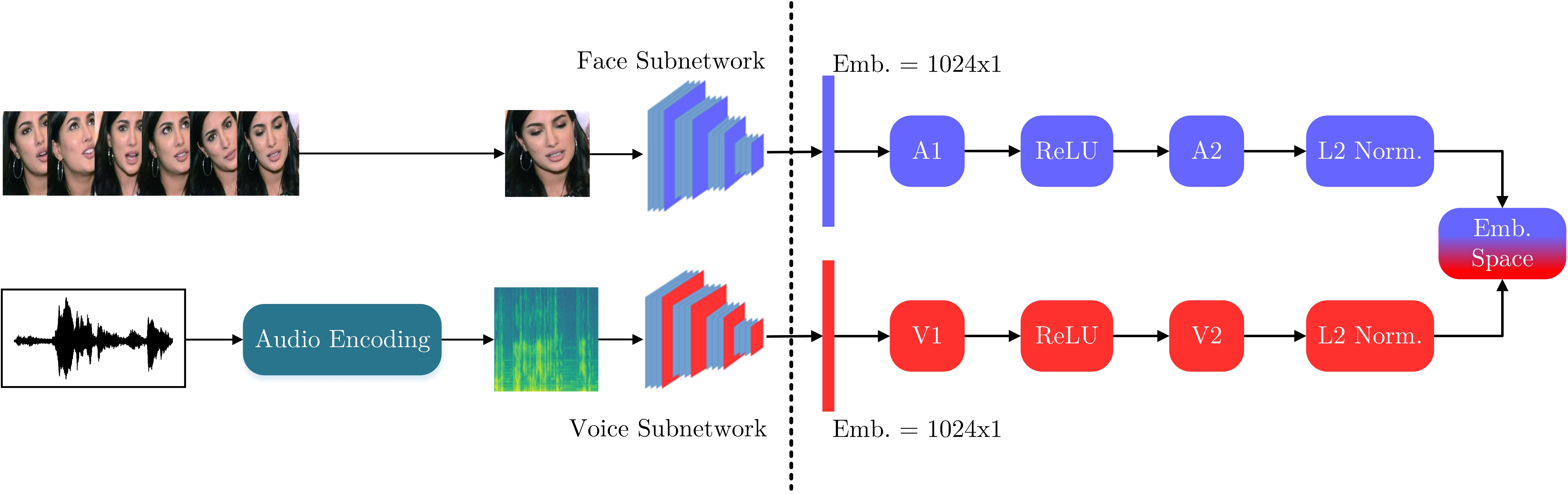}
 \caption{Cross-modal verification network configuration. The left side represents audio and face sub networks trained separately. Afterwards, audio and face embedding is extracted and fed to train a shallow architecture represented on the right side.} 
 \label{fig:cross-modal}
 \end{figure*}

\section{Face-voice Association}
\label{sec:first}
We introduced a cross-modal verification approach to analyze face-voice association across multiple languages using~\textit{MAV-Celeb} dataset in order to answer the question: \\ \\
\textit{Q1. Is face-voice association language independent?}
\\ \\
For example, consider a model trained with faces and voice samples of one language. At inference time, the model is evaluated with faces and audio samples of same language and a completely \textit{unheard} language. This experimental setup provides a foundation to analyze association between faces and voices across languages to answer Q1 with a cross-modal verification method. Therefore, we extract face and voice embedding from two subnetworks trained on VGGFace~\cite{parkhi2015deep} and voice samples from~\textit{MAV-Celeb} dataset respectively. Previous works showed that the faces and voices subnetworks can be trained jointly to bridge the gap between the two~\cite{kim2018learning,nagrani2018learnable}.
However, we built a shallow architecture on top of face and voice embedding to reduce the gap between them to establish baseline on~\textit{MAV-Celeb} dataset. The approach is inspired from the previous work on images and text~\cite{wang2016learning}. The details of these sub networks and shallow architecture are as follow:\\  
\textbf{Face Subnetwork~}--~The face subnetwork is implemented using the VGG-Face CNN descriptor~\cite{parkhi2015deep}. The input to the face subnetwork is an RGB image, cropped from the source frame to include only the face region and resized to $256\times256$. The final fully connected layer of the network produce embedding for every face input.\\
\textbf{Voice Subnetwork~}--~Nagrani et al.~\cite{Nagrani17} introduced VGG-Vox network to process audio information. The network is trained with `softmax' loss function in a typical classification scenario. In the current work, we configure the network to produce  embedding from the \textit{fc7} layer.\\
\textbf{Cross-modal Verification~}--~Finally, we learn a face-voice association for 
cross-modal verification approach using a two stream neural network with two layers of nonlinearities on top of the face and voice embedding.
Fig.~\ref{fig:cross-modal} shows the Two-Branch shallow architecture along with the pre-trained subnetworks.  
The shallow architecture consists of two branches, each composed of fully connected layer with weight matrices $A1$, $V1$ and $A2$, $V2$. In addition, layers are separated by Rectified ($ReLU$) followed by $L2$ normalization.\\
\textbf{Loss Function~}--~Given a training face $f_i$, let  $Y_i^+$  and $Y_i^-$ represent sets of positive and negative voice samples respectively. 
We impose the distance between $f_i$ and each positive voice sample  $y_j$  to be smaller than the distance between $f_i$ and each negative voice sample $y_k$ with margin $m$:
\begin{equation}
\label{eq:sent-pos}
d\left(f_{i}, y_{j}\right)+m<d\left(f_{i}, y_{k}\right) \quad \forall y_{j} \in Y_{i}^{+}, \forall y_{k} \in Y_{i}^{-}.
\end{equation}
Eq.~(\ref{eq:sent-pos}) is modified for a voice $y_{i^{\prime}}$:
\begin{equation}
d\left(f_{j^{\prime}}, y_{i^{\prime}}\right)+m<d\left(f_{k^{\prime}}, y_{i^{\prime}}\right) \quad \forall f_{j^{\prime}} \in X_{i^{\prime}}^{+}, \forall f_{k^{\prime}} \in X_{i^{\prime}}^{-},
\end{equation}
where $X_{i^{\prime}}^{+}$ and $X_{i^{\prime}}^{-}$ represents the sets of positive and negative face for  $y_{i^{\prime}}$.

Finally, constraints are converted to the training objective using hinge loss. The resulting loss function is given by:

\begin{equation}
\begin{aligned}
L(X, Y) &=\sum_{i, j, k} \max \left[0, m+d\left(f_{i}, y_{j}\right)-d\left(f_{i}, y_{k}\right)\right] \\
&+\lambda_{1} \sum_{i^{\prime}, j^{\prime}, k^{\prime}} \max \left[0, m+d\left(f_{j^{\prime}}, y_{i^{\prime}}\right)-d\left(f_{k^{\prime}}, y_{i^{\prime}}\right)\right] \\
&+\lambda_{2} \sum_{i, j, k} \max \left[0, m+d\left(f_{i}, x_{j}\right)-d\left(f_{i}, x_{k}\right)\right] \\
&+\lambda_{3} \sum_{i^{\prime}, j^{\prime}, k^{\prime}} \max \left[0, m+d\left(y_{i^{\prime}}, y_{j^{\prime}}\right)-d\left(y_{i^{\prime}}, y_{k^{\prime}}\right)\right].
\end{aligned}
\end{equation}

The shallow architecture configured with the loss function produce joint embedding of face and voice to study face-voice association across multiple languages using the proposed dataset.
The hyperparameter $\lambda_{1}$ is fixed to $2$. Similarly, $\lambda_{2}$ and $\lambda_{3}$ controls the neighborhood constraint and values are set to $0.1$ or $0.2$ respectively~\cite{wang2016learning}. The distance $d$ is fixed to be the Euclidean distance. 
In addition, triplets are selected within the mini-batch only.

\subsection{Experimental Protocol}
\label{subsec:eval}
We propose an evaluation protocol for a cross-modal verification  method in order to answer Q1. The aim of cross-modal verification task is to verify if an audio sample and a face image belong to the same identity or not based on a threshold value. We report performance on a standard verification metric i.e. Equal Error Rate (EER).

\begin{figure*}[t]
     \centering
     \includegraphics[scale=0.05]{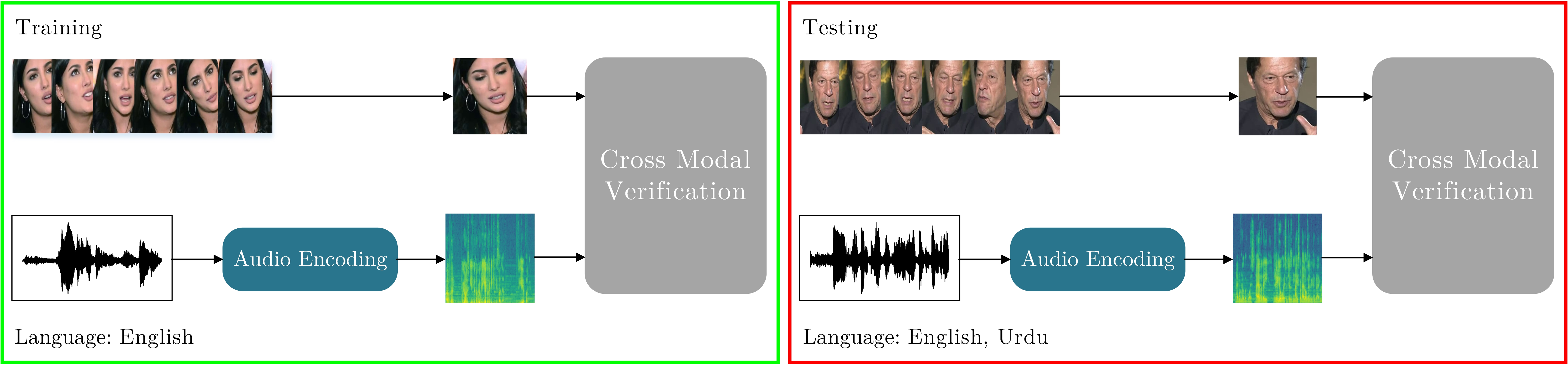}
     \caption{Evaluation protocol to analyze the impact of multiple languages on association between faces and voices.~\textcolor{green}{\textbf{Green}} and the~\textcolor{red}{\textbf{red}} blocks represent training and testing strategies. At test time, the network is evaluated on \textit{unseen-unheard} configuration from the same language (English) \textit{heard} during training along with a completely \textit{unheard} language (Urdu).}
     \label{fig:evaluation-proto}
 \end{figure*}

The~\textit{MAV-Celeb} dataset is divided into train and test splits consisting of disjoint identities from the same language typically known as \textit{unseen-unheard} configuration~\cite{nagrani2018learnable,nagrani2018seeing}. Fig.~\ref{fig:evaluation-proto} shows evaluation protocol during training and testing stages. At inference time, the network is evaluated on a \textit{heard} and completely \textit{unheard} language. The protocol is more challenging than previously known \textit{unseen-unheard} configuration due to the presence of an \textit{unheard} language in addition to disjoint identities.
The dataset splits EU, EH contains $64$--$6$, $78$--$6$ identities for train and test respectively.

\subsection{Experiments and Results}
In first set of experiments, we compare the performance of the proposed Two-Branch network on a cross-modal verification application with previous state-of-the-art between faces and voices ~\cite{nagrani2018learnable,nawaz2019deep,wen2018disjoint}. We extracted face and voice embedding from pretrained VGG-Face CNN descriptor and VoxCeleb respectively. Finally, we train the Two-Branch network for a cross-modal verification application on top of face and voice embedding.
Table~\ref{tab:voxceleb} shows the result along with previous state-of-the-art methods. It is clear that the performance of our method is comparative with state-of-the-art methods, therefore we configure the approach to evaluate cross-modal verification method across multiple languages on ~\textit{MAV-Celeb} dataset to establish baseline results.

\setlength{\tabcolsep}{4pt}
\begin{table}
\begin{center}

\begin{tabular}{|lc|}
\hline
Method $\qquad\qquad$& EER \\
\hline\hline
Two-Branch (Proposed)  & 29.0\\
Learnable Pins-Scratch~\cite{nagrani2018learnable}  & 39.2\\
Learnable Pins-Pretrain~\cite{nagrani2018learnable} & 29.6\\
Single Stream Network~\cite{nawaz2019deep} & 29.5\\
DIMNet~\cite{wen2018disjoint}                       & \textbf{24.5}\\
\hline
\end{tabular}
\end{center}
\caption{Cross-modal verification results with \textit{unseen-unheard} configuration on VoxCeleb dataset. \textbf{(EER: lower is better)}}
\label{tab:voxceleb}
\end{table}
\setlength{\tabcolsep}{1.4pt}

In the second set of experiments, we evaluate Two-Branch network on cross-modal verification method between faces and voices to measure performance on \textit{heard} and \textit{unheard} configurations of~\textit{MAV-Celeb} dataset. 
Table~\ref{tab:cross-modal} shows the result of face-voice association across multiple languages using the proposed evaluation protocol. We observed performance drop across $2$ splits, which clearly demonstrate that the association between faces and voices is not language independent.  We observed that the performance degradation is due to different data distributions of the two languages, typically known as domain shift~\cite{shimodaira2000improving}. Moreover, the model does not generalize well to other \textit{unheard} language. However, the performance is still better than random verification, which is not trivial considering the challenging nature and configuration of the proposed evaluation protocol.

\setlength{\tabcolsep}{4pt}
\begin{table}[t]
\begin{center}
\resizebox{0.48\textwidth}{!}{%
\begin{tabular}{|llcc|}
\hline
       &  & \multicolumn{2}{c|}{\textbf{EU}} \\
\hline\hline
Method & Configuration & Eng. test  & Urdu test   \\
             & & (EER)                          & (EER) \\

\hline
\multirow{2}{*}{Two-Branch (Proposed)} & Eng. train     & 45.1    & 48.3$\downarrow$\tiny7.1    \\
                                       & Urdu train     & 47.0$\downarrow$\tiny6.3   & 44.3    \\

\hline\hline
       &  & \multicolumn{2}{c|}{\textbf{EH}} \\      
\hline\hline
 &  & Eng. test & Hindi test  \\
             & & (EER)                          & (EER) \\
\hline
\multirow{2}{*}{Two-Branch (Proposed)} & Eng. train  & 35.7  & 36.7$\downarrow$\tiny2.8 \\
                                       & Hindi train    & 38.9$\downarrow$\tiny4.2  & 37.3  \\
\hline
\end{tabular}}
\end{center}
\caption{Cross-modal verification between face and voice across multiple language on various test configurations of \textit{MAV-Celeb} dataset. The down arrow($\downarrow$) represents percentage decrease in performance. \textbf{(EER: lower is better)}}
\label{tab:cross-modal}
\end{table}
\setlength{\tabcolsep}{1.4pt}

\section{Speaker Recognition}
\label{sec:second}
This section investigates the performance of speaker recognition across multiple languages to answer the following question. 
\\ \\
\textit{Q2. Can a speaker be recognised irrespective of the spoken language?} 
\\ \\
For example, consider a model trained with voice samples of one language. At inference time, the model is evaluated with audio samples of the same language and a completely \textit{unheard} language of the same speaker. Therefore, the experimental setup provides a foundation for speaker recognition across multiple languages to answer Q2. 

 \begin{figure*}[t]
     \centering
     \includegraphics[scale=0.055]{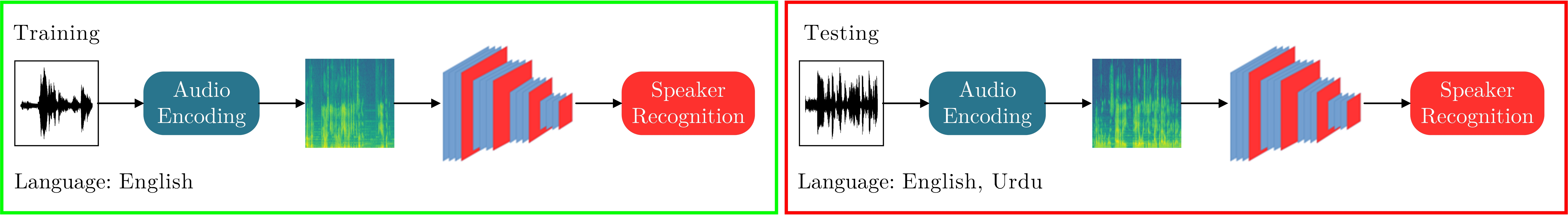}
     \caption{Evaluation protocol to analyze the impact of multiple languages on speaker recognition.~\textcolor{green}{\textbf{Green}} and the~\textcolor{red}{\textbf{red}} blocks represent training and testing strategies respectively. At test time, the network is evaluated on the same language \textit{heard} during training along with completely \textit{unheard} language of the same identities.}
     \label{fig:spprotocol}
 \end{figure*}

\subsection{Baselines}
We employed following $3$ methods to establish baseline results for speaker recognition across multiple languages using \textit{MAV-Celeb} dataset to answer Q2. \\
\textbf{VGG-Vox~}--~ Nagrani et al.~\cite{Nagrani17} introduced VGG-Vox network by modifying VGG-M~\cite{chatfield2014return} model to adapt to the spectrogram input. Specifically, the fully connected \textit{fc6} layer of VGG-M is replaced by two layers – a fully connected layer and an average pool layer. \\
\textbf{Utterance Level~}--~ Xie et al.~\cite{xie2019utterance} presented a deep neural network based on 
NetVLAD or GhostVLAD layer that is used to aggregate thin-ResNet architecture frame features. \\
\textbf{SincNet~}--~ Ravanelli et al.~\cite{ravanelli2018speaker} presented a deep neural model to process raw audio samples and learn features. The approach is based on parameterized sinc function for band-pass filtering that is used to convolve the wavefor to extract basic low-level features to be later processed by the deeper layers of the network.


\subsection{Experimental Protocol}
We proposed an evaluation protocol in order to analyze the impact of multiple languages on speaker recognition to answer Q2. The~\textit{MAV-Celeb} dataset is divided into typical classification scenario for speaker identification. However, different voice tracks of the same person are used for train, validation and test~\cite{Nagrani17}.
The network is trained with one language and tested with the same language and a completely \textit{unheard} language of same identities.  Moreover, the dataset is split into disjoint identities for speaker verification~\cite{Nagrani17}. 
Fig.~\ref{fig:spprotocol} shows evaluation protocol for speaker recognition across multiple languages. The protocol is consistent with previous studies on human subjects for speaker identification~\cite{perrachione2011human}.
For identification and verification, we employed Top-1 accuracy and EER metrics to report performance.

\setlength{\tabcolsep}{4pt}
\begin{table}[t]
\begin{center}
\resizebox{0.48\textwidth}{!}{%
\begin{tabular}{|llcc|}
\hline\hline
       &  & \multicolumn{2}{c|}{\textbf{EU}} \\
\hline
Method & Configuration & Eng. test  & Urdu test \\
             & & (Top-1 )                          & (Top-1) \\
\hline
\multirow{2}{*}{VGGVox-Scratch}      & Eng. train     & 56.1    & 37.7$\downarrow$\tiny32.8	   \\
                                     & Urdu train     & 45.4$\downarrow$\tiny19.9    & 56.7   \\
                                     
\hline
\multirow{2}{*}{VGGVox-Pretrain(VoxCeleb1)} & Eng. train     & 41.0    & 38.0$\downarrow$\tiny7.0  \\
                                            & Urdu train     & 46.0$\uparrow$\tiny6.8    & 43.0  \\
\hline
\multirow{2}{*}{SincNet}   & Eng. train     & 32.5  & 19.3$\downarrow$\tiny40.6    \\
                                            & Urdu train     & 21.3$\downarrow$\tiny47.0    & 40.2 \\

\hline\hline
       &  & \multicolumn{2}{c|}{\textbf{EH}} \\      
\hline\hline
 &  & Eng. test & Hindi test  \\
             & & (Top-1\%)                          & (Top-1\%) \\
\hline
\multirow{2}{*}{VGGVox-Scratch}      & Eng. train     & 60.2    & 55.0$\downarrow$\tiny8.6     \\
                                     & Hindi train    & 47.5$\downarrow$\tiny13.6     & 54.7  \\
                                     
\hline
\multirow{2}{*}{VGGVox-Pretrain(VoxCeleb1)} & Eng. train     & 43.0    & 32.0$\downarrow$\tiny25.5   \\
                                            & Hindi train     & 43.0$\downarrow$\tiny10.4     & 48.0  \\
\hline
\multirow{2}{*}{SincNet}   & Eng. train     & 23.9  & 8.7$\downarrow$\tiny63.6   \\
                           & Hindi train     & 14.4 $\downarrow$\tiny43.0  & 25.3 \\

\hline
\end{tabular}}
\end{center}
\caption{Speaker identification results across multiple languages on test configurations of \textit{MAV-Celeb} dataset. The down arrow($\downarrow$)  and up arrow($\uparrow$) represents percentage decrease and increase in performance.  (\textbf{Top-1: higher is better})}
\label{tab:identi}
\end{table}
\setlength{\tabcolsep}{1.4pt}

\subsection{Experiments and Results}

We evaluate the performance of speaker recognition across multiple languages on $3$ baseline methods. Table~\ref{tab:identi} shows speaker identification performance on $2$ splits (EU, EH) of \textit{MAV-Celeb} dataset. We note that the performance drop occurred on a completely \textit{unheard} language  across all baseline methods for both splits. The speaker identification models (VGG-Vox, SincNet) do not generalize well on \textit{unheard} language and is overfitted on a particular language. However, its performance is quantitatively better than random classification on \textit{unheard} language.
Based on these results, we conclude that speaker identification is a language dependent task.  
Furthermore, these results are inline with the previous studies which show that human's speaker identification performance is higher on people speaking familiar language than people speaking \textit{unknown} language~\cite{perrachione2011human}.

Similarly, Table~\ref{tab:verification} shows speaker verification performance on $2$ splits (EU, EH) of \textit{MAV-Celeb} dataset.  We note that performance drop occurred on a completely \textit{unheard} language for EU and EH across three baseline methods. Therefore, speaker verification is also not language independent.

\setlength{\tabcolsep}{4pt}
\begin{table}[t]
\begin{center}
\resizebox{0.48\textwidth}{!}{%
\begin{tabular}{|llcc|}
\hline
       &  & \multicolumn{2}{c|}{\textbf{EU}} \\
\hline\hline
Method & Configuration & Eng. test  & Urdu test  \\
             & & (EER)                          & (EER) \\
\hline
\multirow{2}{*}{VVGVox-Scratch}      & Eng. train     & 36.5    & 41.5$\downarrow$\tiny13.7 \\
                                     & Urdu train     & 40.3$\downarrow$\tiny3.0     & 39.1   \\
\hline
\multirow{2}{*}{Utterance Level-Scratch}   & Eng. train   & 39.9   & 45.5$\downarrow$\tiny14.0     \\
                                           & Urdu train   & 42.5$\downarrow$\tiny10.3 & 38.5  \\

\hline\hline
       &  & \multicolumn{2}{c|}{\textbf{EH}} \\      
\hline\hline
 &  & Eng. test & Hindi test  \\
             & & (EER)                          & (EER) \\
\hline
\multirow{2}{*}{VVGVox-Scratch}  & Eng. train     & 29.6    & 37.8$\downarrow$\tiny27.7  \\
                                 & Hindi train     & 32.7$\downarrow$\tiny15.9     & 28.2 \\
\hline
\multirow{2}{*}{Utterance Level-Scratch}   & Eng. train    & 34.9   & 40.6$\downarrow$\tiny 16.3  \\
                                           & Hindi train   & 42.7$\downarrow$\tiny 18.9    & 35.9  \\

\hline
\end{tabular}}
\end{center}
\caption{Speaker verification results across multiple languages on various test configurations of \textit{MAV-Celeb} dataset. The down arrow($\downarrow$) represents percentage decrease in performance.  (\textbf{EER: lower is better})}
\label{tab:verification}
\end{table}
\setlength{\tabcolsep}{1.4pt}

 \begin{figure*}
     \centering
     \includegraphics[scale=0.60]{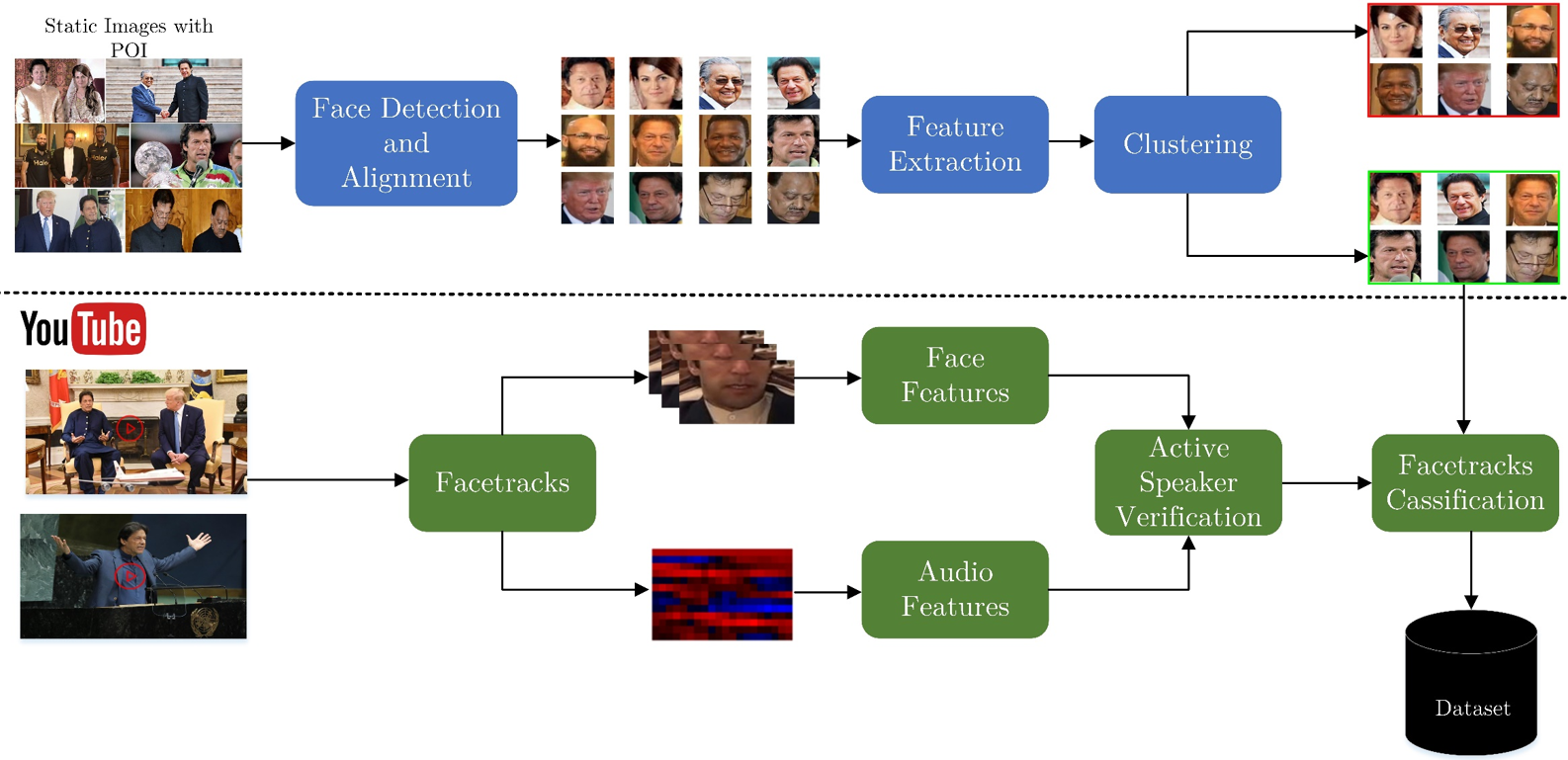}
     \caption{Data collection pipeline. It consists of two blocks with upper block download static images  while the bottom block download and process videos from YouTube.} 
     \label{fig:data-collection}
 \end{figure*}

\section{Conclusion}
\label{sec:con}
In this work, effect of language is explored on cross-modal verification between faces and voices along with speaker recognition tasks.  A new audio-visual dataset consisting of $154$ celebrities is presented with language level annotation. The dataset contains $2$ splits having same set of identities speaking English/Urdu and English/Hindi. In the cross-modal verification experiment by changing training and test language, performance drop is observed indicating that face-association is not language independent. In case of speaker recognition, similar drop in performance is observed, thus concludes that speaker recognition is also language dependent task. The reason in performance is due to the domain shift caused by two different languages. 

\appendix

\section{Dataset Collection Pipeline}
\label{append:data-pipeline}
In this section we present a semi-automated  pipeline inspired by Nagrani et al.~\cite{Nagrani17} for collecting the proposed dataset. 
The pipeline is shown in Fig.~\ref{fig:data-collection} and various stages are discussed below. \\
{\bf Stage 1 -- List of Persons of Interest:} 
In this stage, candidate list of Persons of Interest (POIs) is generated by scraping Wikipedia. The POIs cover over a wide range of identities including sports persons, actors, actresses, politicians, entrepreneurs and singers. \\
{\bf Stage 2 -- Collecting list of YouTube links:} 
In this step we used crowd-sourcing to collect lists of YouTube videos. Keywords like ``Urdu interview'', ``English Interview'',``public speech English'', ``public speech Urdu'' are appended to increase the likelihood that search results contain an instance of POI speaking. The links of search results are stored in text files. Videos are then automatically downloaded using the links from these text files.\\
{\bf Stage 3 -- Face tracks:} 
In this stage, we employed joint face detection and alignment using Multi-task Cascaded Convolutional Networks (MTCNN) for face detection and alignment~\cite{zhang2016joint}. MTCNN can detect faces in extreme conditions, and different poses. After face detection and alignment, shot boundaries are detected by comparing color histograms across consecutive frames. Based on key frames from shot boundaries and detected faces, face tracks are generated. \\
{\bf Stage 4 -- Active speaker verification:} 
The goal of this stage is to determine the visible speaking faces. We carried out this stage by using ‘SyncNet’ which estimates the correlation between mouth motion and audio tracks~\cite{chung2016out}. Based on scores from this model, face tracks with no visible speaking faces, voice-over and background speech are rejected. \\
{\bf Stage 5 -- Static Images:} 
In this stage, static images are automatically downloaded using Google Custom Search API based on list of POIs obtained from stage $1$. 
MTCNN is employed to detect and align static face images. 
A clustering mechanism based on a popular density-based clustering algorithm DBSCAN~\cite{ester1996density} is used to remove false positives from the detected and aligned faces. Interestingly, DBSCAN does not require a priori specification of the number of clusters in the data. Intuitively, the clustering algorithm groups faces of an identity that are closely packed together.\\
{\bf Stage 6 -- Face tracks classification:} 
In this stage, active speaker face tracks are classified if they belong to POI or not. 
We trained an Inception ResNet V1 network~\cite{szegedy2017inception} on VGGFace2 dataset~\cite{cao2018vggface2} with center loss~\cite{wen2016discriminative} to extract discriminative embedding from face tracks and static images. 
A classifier is trained based on Support Vector Machine with static face embedding. Finally, classification is performed using a score with a threshold obtained from each face track.

{\small
\bibliographystyle{ieee_fullname}
\bibliography{egbib}
}

\end{document}